\documentclass[11pt]{article}

\usepackage[final]{acl}

\usepackage{times}
\usepackage{latexsym}

\usepackage[T1]{fontenc}
\usepackage[utf8]{inputenc}

\usepackage{microtype}

\usepackage{inconsolata}

\usepackage{graphicx}

\title{Clinical named entity recognition in the Portuguese language: a benchmark of modern BERT models and LLMs}

\author{
\textbf{Vinicius A. de Almeida\textsuperscript{1,2}} \quad
\textbf{Sandro S. da Silva\textsuperscript{1}} \quad
\textbf{Josimar Chire\textsuperscript{1}} \quad
\textbf{Nícolas H. Borges\textsuperscript{1,3}} \quad \\
[0.4em]
\textbf{Leonardo Vicenzi\textsuperscript{1}} \quad
\textbf{Helena Kociolek\textsuperscript{1,3,5}} \quad
\textbf{Sarah M. de C. Rocha\textsuperscript{1,3}} \quad
\textbf{Frederico N. Gomes\textsuperscript{1,3}} \\
[0.4em]
\textbf{Júlia C. Ferreira\textsuperscript{1,3}} \quad
\textbf{Oge Marques\textsuperscript{4}} \quad
\textbf{Lucas E. S. e Oliveira\textsuperscript{1,3}} \\
[0.8em]
\textsuperscript{1}Spesia, Curitiba - PR, Brazil \\
\textsuperscript{2}Faculdade de Medicina, Universidade de São Paulo, São Paulo - SP, Brazil \\
\textsuperscript{3}Pontifícia Universidade Católica do Paraná (PUCPR), Curitiba - PR, Brazil \\
\textsuperscript{4}Florida Atlantic University, Boca Raton - FL, USA \\
\textsuperscript{5}Universidade Federal do Paraná (UFPR), Curitiba - PR, Brazil \\
\small \textbf{Correspondence:} \href{mailto:vinicius.almeida@spesia.com.br}{vinicius.almeida@spesia.com.br}
}

\begin{document}
\maketitle
\begin{abstract}
\textbf{Background}: Clinical notes hold valuable unstructured information. Named entity recognition (NER) enables the automatic extraction of medical concepts; however, benchmarks for Portuguese remain scarce.

\textbf{Objectives}: To evaluate BERT models and large language models (LLMs) for clinical NER in Portuguese and test strategies to address multilabel imbalance.

\textbf{Methods}: We compared BioBERTpt, BERTimbau, ModernBERT, and mmBERT with LLMs such as GPT-5 and Gemini-2.5, using the public SemClinBr corpus and a private breast-cancer dataset. Models were trained under identical conditions and evaluated with precision, recall, and F1-scores. Iterative stratification, weighted loss, and oversampling were tested to handle class imbalance.

\textbf{Results}: mmBERT-base achieved the best results (micro F1 = 0.76), outperforming all models. Iterative stratification improved class balance and overall performance.

\textbf{Conclusions}: Multilingual BERT models, especially mmBERT, perform strongly for Portuguese clinical NER and can run locally with limited resources. Balanced data-splitting strategies further enhance results.
\end{abstract}

\section{Introduction}

Unstructured text data is ubiquitous in healthcare \cite{Sedlakova2023, Nesca2022-cw}. Patient complaints, clinical findings, relevant laboratory results, care plans, and the healthcare professional rationale for their decisions are essential information often lost in electronic health record notes. Natural language processing techniques, particularly with the current advancements, show potential for healthcare institutions to leverage text data for management, research, and quality of care improvement \cite{ALAFARI2025100725}.

Recent advancements in language modeling, especially with large language models (LLMs), have motivated research in several NLP tasks, such as named-entity extraction \cite{KOCAMAN2022100373, monajatipoor2024llmsbiomedicinestudyclinical}, information retrieval \cite{Liu2025}, and question-answering \cite{Kell2024}. This momentum has fueled the development of specialized encoders tailored to specific languages and domains.

In the Portuguese context, BERT-based models have been adapted for complex legislative jargon \cite{nunes-etal-2024-named}, applied to financial Named Entity Recognition \cite{abilio_evaluating_2024}, and subjected to comparative evaluations across general and specific domains \cite{bracis}.

With respect to named entity recognition in the clinical context, modern BERT models have emerged; however, they have not yet been evaluated on clinical datasets in the Portuguese language. The \texttt{mmBERT} models \cite{mmbert} in particular demonstrated improved multilingual capabilities that may benefit Portuguese clinical NLP tasks. 

Finally, the results found in benchmarks using public datasets do not always translate to private data \cite{deng2024investigatingdatacontaminationmodern, dong-etal-2024-generalization}. The reproducibility of results in private data is essential for technology adoption, especially in healthcare.

In this scenario, the objectives of this study are two-fold: to evaluate the performance of BERT models and LLMs in clinical named entity recognition tasks using both public and private datasets, and to explore strategies for handling multilabel token classification datasets with severely imbalanced classes.

\section{Methods}

\subsection{Datasets}

Two datasets were used in this benchmark: the SemClinBr dataset and a breast cancer clinical notes dataset.

\subsubsection{SemClinBr}

The SemClinBr \cite{Oliveira2022} corpus is a publicly available, semantically annotated dataset of clinical notes from more than twelve medical specialties, encompassing both inpatient and outpatient scenarios. It contains 1,000 entries, including discharge summaries, admission notes, nursing notes, and ambulatory notes. The annotation protocol for named-entity recognition tasks included 15 UMLS \cite{Humphreys1993} semantic groups, 12 of which contained at least 30 entities in the dataset. The quantity of clinical records containing each semantic group, together with examples, is available in \textbf{Table \ref{tab:semclinbr}}.

\subsubsection{Breast cancer clinical notes}

A restricted-access dataset of clinical notes was organized with a sample from a private oncological healthcare center in the southern region of Brazil. Five hundred ambulatory visit notes from breast cancer patients were sampled from the center's database. 

A team of five medical students, coordinated by one medical doctor, was responsible for annotating entities and resolving disagreements until consensus was reached. The annotation team held regular meetings to address doubts and inconsistencies in the annotation process.

Ten entities were annotated based on their relevance to the care process of breast cancer patients: estrogen receptor (ER), progesterone receptor (PR), human epidermal growth factor receptor 2 (HER2), histological type, surgical procedure, menopause status, presence of metastasis, metastasis location, breast cancer susceptibility gene (BRCA), and results of FISH (fluorescence in situ hybridization), CISH (chromogenic in situ hybridization), or SISH (silver in situ hybridization) tests. The quantity of clinical records containing each entity and an annotated example of each one is available in \textbf{Table \ref{tab:breastcancerdataset}}.

% SemClinBr table
\begin{table*}
    \centering
    \begin{tabular}{lcl}
        \hline
        \textbf{Entity} & \textbf{Number of records} & \textbf{Examples} \\       
        \hline
        Procedures & 984 & `medicada', `soroterapia', `mudança de decúbito' \\
        Disorders & 979 & `sonolenta', `dependente', `diabetes' \\
        Concepts \& Ideas & 887 & `2 dias', `2 anos' \\
        Living Beings & 715 & `filho', `neto' \\
        Anatomy & 705 & `calcâneo', `pescoço', `perna E' \\
        Chemicals \& Drugs & 601 & `tazocin', `ATB', `cálcio' \\
        Physiology & 566 & `evacuação', `diurese' \\
        Phenomena & 492 & `dextro de 88mg/dl' \\
        Devices & 416 & `acesso venoso periférico' \\
        Organizations & 348 & `setor de radiografia', `hospital' \\
        Objects & 122 & `leito', `óculos', `próteses dentárias' \\
        Activities \& Behaviors & 36 & `fuma', `empregada doméstica', `sedentarismo' \\
        \hline
    \end{tabular}
    \caption{Number of records per entity in the SemClinBr dataset and entity examples. The UMLS semantic groups `Genes \& Molecular Sequences', `Occupations', and `Geographic Areas' were not included in this study due to the small number of examples in the dataset.}
    \label{tab:semclinbr}
\end{table*}

% Doccano table
\begin{table*}
\centering
    \begin{tabular}{lcl}
    \hline
    \textbf{Entity}                  & \textbf{Number of records} & \textbf{Example (entity in bold)}\\       
    \hline
    ER                      &     346      &   `IHQ: \textbf{RE-}/RP-/Her2 3+/ gata3 +++'                  \\       
    HER2                    &     343      &   `IHQ: RE-/RP-/\textbf{Her2 3+}/ gata3 +++'\\           
    Histological type       &     336      &   `\textbf{adenocarcinoma pouco diferenciado} associado a...'\\           
    PR                      &     335      &  `IHQ: RE-/\textbf{RP-}/Her2 3+/ gata3 +++' \\
    Surgery                 &     296      &  `\textbf{Mastectomia mama dir} + \textbf{EA} em 2011'                   \\
    Menopause status        &     212      &  `Paciente de 51 anos, \textbf{pré-menopausa}'  \\
    Metastasis location     &     117      &  `Ca de mama           progressão de doença para \textbf{ossos}'    \\
    Distant metastasis      &     72      &   `\textbf{nódulos peritoneais e lesão na bacia metastástica}'                  \\
    BRCA                    &     62      &  `Análises genéticas: \textbf{BRCA 1 e 2 não mutados}'     \\
    FISH/CISH/SISH          &     32      &   `\textbf{FISH negativo}'  \\
    \hline
    \end{tabular}
    \caption{Number of records per entity in the breast cancer clinical notes dataset and entity examples. ER, estrogen receptor; HER2, human epidermal growth factor receptor 2; PR, progesterone receptor; BRCA, breast cancer susceptibility gene; FISH, fluorescence in situ hybridization; CISH, chromogenic in situ hybridization; SISH, silver in situ hybridization.}
    \label{tab:breastcancerdataset}
\end{table*}

\subsection{Metrics}

Models were evaluated using the F1-score, precision, and recall, aggregated across all labels with both micro and macro averages. These metrics were computed using the scikit-learn Python library \cite{scikitlearnSklearnmetrics}.

In order to compute these metrics, a probability threshold must be defined for each model. The threshold that maximized each model's F1-score was chosen for the sake of comparison. 

In addition, precision-recall curves were plotted for each model and label to better describe the precision-recall trade-off.

\subsection{Base models}

Bidirectional Encoder Representations from Transformers (BERT) \cite{bert} is a transformer-based neural network architecture for language modeling. It has demonstrated state-of-the-art performance in many natural language processing (NLP) tasks, including named entity recognition (NER) in the clinical domain in multiple languages \cite{schneider-etal-2020-biobertpt, danu2025multilingualclinicalnerdiseases, 10593504}.

\textbf{BioBERTpt}. The BioBERTpt models \cite{schneider-etal-2020-biobertpt} were built on top of the original multilingual BERT models from Google \cite{bert}. Three versions were released: \texttt{biobertpt-clin}, trained on 2 million Brazilian Portuguese clinical notes from brazilian hospitals; \texttt{biobertpt-bio}, trained on 16.4 million words from scientific papers from Pubmed \cite{nihPubMed} and Scielo \cite{scielo}; and \texttt{biobertpt-all}, trained with a combination of both corpora. To the best of the authors knowledge, it holds state-of-the-art performance on the SemClinBr dataset.

\textbf{ModernBERT}. The ModernBERT models \cite{modernbert} incorporate many recent optimizations of language modeling in an encoder-only architecture. These optimizations include both architectural improvements (e.g., disabling bias terms in linear layers, adopting rotary positional embeddings, and using pre-normalization) and efficiency improvements (e.g., alternating attention, unpadding during training and inference, and flash attention). This strategy demonstrated significant improvements in the context window, natural language understanding, and information retrieval tasks. 

\textbf{mmBERT}. The mmBERT models \cite{mmbert} were built with an architecture inspired by ModernBERT \cite{modernbert} and a novel pre-training protocol aimed at improving multilingual capabilities. The pre-training involved three phases in which low-resource language data was gradually included, and the ratio between high-, mid-, and low-resource languages was dynamically adjusted. The authors demonstrated improved performance in NLP tasks when compared to similar multilingual models, including retrieval, reranking, and summarization.

\textbf{BERTimbau}. The BERTimbau models \cite{bertimbau} build upon the same architecture as the original BERT models \cite{bert}. They were trained on 2.68 billion tokens from the BRWAC corpus, the largest Portuguese open corpus up to that time. They also demonstrated strong results in Portuguese NER and text similarity tasks.

This study also compared the performance of BERT models to large language models (LLMs) prompted with a few-shot strategy. The prompt template is available in \textbf{Table \ref{tab:prompt_strategy}}. The models included in this baseline were: \texttt{gemini-2.5-flash}, \texttt{gemini-2.5-flash-lite}, \texttt{gemini-2.5-pro}, \texttt{gpt-4.1}, \texttt{gpt-5-mini}, \texttt{gpt-5}, and \texttt{gpt-5-nano}. All \texttt{gpt-5} models were prompted with the minimal reasoning effort option to control costs and reduce inference time. Only the \texttt{gpt-5} model was tested on `low', `medium' and `high' reasoning efforts. LLMs were evaluated using only the SemClinBr dataset to preserve the privacy of the restricted-access breast cancer dataset.

\begin{table}
\centering
\begin{tabular}{p{1cm}p{6cm}}
\hline
\textbf{Role} & \textbf{Prompt} \\ 
\hline
\texttt{system} &
\textit{Extract the following entities from the given clinical note in order of appearance. 
Use exact text for extractions. Do not paraphrase or overlap entities.
The entities to extract are:} \texttt{\{entities\}}. 
\textit{See the examples below for the expected format.}

\# Clinical note example: \texttt{\{example\_text\}}

\# Ideal response example: \texttt{\{example\_response\}} \\ [6pt]
\texttt{user} &
\# Clinical note: \texttt{\{input\_text\}} \\
\hline
\end{tabular}
\caption{Prompt strategy used for entity extraction in clinical notes using LLMs. The fields defined with curly brackets were replaced by the respective entities, text and JSON object expected as the response.}
\label{tab:prompt_strategy}
\end{table}

\subsection{Training setup}

NER was framed in this study as a multilabel token classification task. Each dataset was split into training, validation, and test datasets, following the proportions of 60\%, 20\%, and 20\%. Two different techniques for splitting data were tested and compared: simple randomization and iterative data stratification. 

Multilabel datasets often suffer from severely imbalanced classes, which diminishes the capabilities of any algorithm that learns from the data. Splitting the data while preserving the representation of rare classes is an important but often overlooked step. The iterative stratification algorithm \cite{sechidis2011stratification, pmlr-v74-szymański17a} prioritizes the distribution of rarer classes while taking into account how a particular class is distributed in combination with other classes. We used the implementation available at scikit-multilearn \cite{scikitScikitmultilearnMultilabel}. 

Two entity annotation schemes were evaluated in this study: \textbf{IO}, in which tokens can receive an inside tag `I' representing a class, or an outside `O' tag representing no class; and \textbf{BIO}, which also includes a beginning `B' tag for each class.

The training of each BERT model was implemented in Python using  the \texttt{transformers} library \cite{huggingfaceTransformers}. For consistency, the same hyperparameters were used whenever possible: 0.00005 as the learning rate, 10 as the batch size, 5 for gradient accumulation, and a maximum input length of 512. The metric used for the best model selection was the micro F1 score on the validation set. Training was stopped early with a patience parameter of 5. The loss function was the binary cross entropy loss. Different training parameters were used in case a model failed to converge. In the breast cancer dataset, a learning rate of 0.0001 and a patience of 15 were used.

In order to account for class imbalance, unweighted and weighted losses were compared. Class weights were computed as suggested in the PyTorch documentation \cite{pytorchBCEWithLogitsLossx2014}. The weight $w_c$ for a class $c$ was defined as $w_c = \frac{N_{c'}}{N_{c}}$, where $N_{c'}$ is the number of clinical notes not containing class $c$ entities, and $N_{c}$ is the number of clinical notes containing class $c$ entities.

Oversampling was also explored to tackle class imbalance. The oversampling algorithm replicated examples in the training dataset, prioritizing the less frequent classes until their frequency was similar to the average class frequency.

For reproducibility, the source code of this study was made available in the GitHub repository: \url{https://github.com/GRUPOMED4U/clinical\_ner\_benchmark\_paper}.

\subsection{Hardware}

This study was conducted on a desktop equipped with a 13th Gen Intel(R) Core(TM) i9-13900K (3.00 GHz, 24 cores), 64 GB DDR5 RAM, and an NVIDIA GeForce RTX 4090 GPU with 24 GB VRAM. The system had a 4 TB Kingston SSD for storage and Windows 11 Pro as the operating system.

\section{Results}

\subsection{NER benchmark}

All BERT models were trained and evaluated on both datasets. The metrics on the test split are presented in \textbf{Table \ref{tab:benchmark}}. The F1-scores reported are the maximum achievable F-score computed with a shared probability threshold across all classes. 

The \texttt{mmBERT} models achieved the best results in both datasets, and \texttt{mmBERT-base} established a new state-of-the-art on the SemClinBr dataset (micro F1-score = 0.7646). BERT models, in general, have demonstrated better performance when compared to leading LLMs in the industry. They also had faster inference and ran locally on domestic hardware, enabling experiments with no privacy risks.

The experiments involving increased reasoning effort using the \texttt{gpt-5} model demonstrate a modest gain in performance that was not able to surpass every BERT model. The increase in performance was followed by a significant increase in cost and inference time, as shown in \textbf{Table \ref{tab:llmsresourceuse}}. The increased resource requirements may be prohibitive, depending on the setting and the application.

\begin{table}
\centering
    \begin{tabular}{lccc}
    \hline
    \textbf{Effort level} & \textbf{Time} & \textbf{Token usage} & \textbf{Cost}\\
    \hline
    minimal & 18.3  & 3064  & 9.57 \\
    low     & 81.4  & 6090  & 39.83 \\
    medium  & 115.5  & 10,508  & 84.01 \\
    \hline
    \end{tabular}
    \caption{Resource usage of the \texttt{gpt-5} model in each reasoning effort level. The measures of time, token usage and cost represent the average across all responses. Time is measured in seconds. Cost is estimated per 1000 responses and is presented in USD considering the pricing at November 14th, 2025. Token usage refers to the total number of tokens (prompt and completion tokens).}
    \label{tab:llmsresourceuse}
\end{table}

\begin{table*}
\centering
    \begin{tabular}{l|cccc|cccc}
    \hline
    \textbf{Model} & \multicolumn{4}{c|}{\textbf{SemClinBr}} & \multicolumn{4}{c}{\textbf{Breast cancer dataset}}\\
    & \multicolumn{2}{c}{\textbf{Macro F1-score}} & \multicolumn{2}{c|}{\textbf{Micro F1-score}} & \multicolumn{2}{c}{\textbf{Macro F1-score}} & \multicolumn{2}{c}{\textbf{Micro F1-score}} \\
    & IO & BIO & IO & BIO & IO & BIO & IO & BIO \\
    \hline
    \textbf{BioBERTpt} & & & & & & & & \\
    biobertpt-all & 0.6443 & 0.6207 & 0.6830 & 0.6573 & 0.7473 & 0.3671 & 0.7602 & 0.6124\\
    biobertpt-clin & 0.6489 & 0.6470 & 0.6981 & 0.6959 & 0.7527 & 0.3982 & 0.7728 & 0.6508 \\
    biobertpt-bio & 0.5839 & 0.5543 & 0.6548 & 0.6208 & 0.7268 & 0.3741 & 0.7458 & 0.6319\\
    & & & & & & & &\\
    \textbf{mmBERT} & & & & & & & & \\
    mmBERT-base & 0.7139 & 0.4231 & 0.7646 & 0.7444 & 0.7445 & 0.4945 & 0.7723 & 0.6598 \\
    mmBERT-small & 0.6667 & 0.3968 & 0.7264 & 0.7055 & 0.7212 & 0.5398 & 0.7558 & 0.6829 \\
    & & & & & & & &\\
    \textbf{ModernBERT} & & & & & & & & \\
    ModernBERT-base & 0.5475 & 0.2277 & 0.6494 & 0.5654 & 0.6987 & 0.3814 & 0.7259 & 0.6796 \\
    ModernBERT-large & 0.5895 & 0.3258 & 0.6987 & 0.6851 & 0.7104 & 0.5056 & 0.7488 & 0.6648 \\
    & & & & & & & &\\
    \textbf{BERTimbau} & & & & & & & & \\
    bert-base-portuguese-cased & 0.5839 & 0.5000 & 0.6592 & 0.6094 & 0.0055 & 0.1220 & 0.3182 & 0.1218 \\
    bert-large-portuguese-cased & 0.5769 & 0.6264 & 0.6421 & 0.6474 & 0.0130 & 0.0038 & 0.0130 & 0.0315 \\
    & & & & & & & &\\
    \textbf{OpenAI {*}} & & & & & & & & \\
    gpt-4.1 &\multicolumn{2}{c}{0.4795}&\multicolumn{2}{c|}{0.5328}& & & & \\
    gpt-5-nano minimal&\multicolumn{2}{c}{0.1870}&\multicolumn{2}{c|}{0.2507}& & & & \\
    gpt-5-mini minimal&\multicolumn{2}{c}{0.4489}&\multicolumn{2}{c|}{0.4746}& & & & \\
    gpt-5 minimal&\multicolumn{2}{c}{0.5878}&\multicolumn{2}{c|}{0.6604}& & & & \\
    gpt-5 low&\multicolumn{2}{c}{0.6311}&\multicolumn{2}{c|}{0.6828}& & & & \\
    gpt-5 medium&\multicolumn{2}{c}{0.6338}&\multicolumn{2}{c|}{0.6907}& & & & \\
    gpt-5 high&\multicolumn{2}{c}{**}&\multicolumn{2}{c|}{**}& & & & \\
    & & & & & & & &\\
    \textbf{Gemini} & & & & & & & & \\
    gemini-2.5-flash-lite &\multicolumn{2}{c}{0.4849}&\multicolumn{2}{c|}{0.5781}& & & & \\
    gemini-2.5-flash &\multicolumn{2}{c}{0.5965}&\multicolumn{2}{c|}{0.6528}& & & & \\
    gemini-2.5-pro &\multicolumn{2}{c}{0.5965}&\multicolumn{2}{c|}{0.6528}& & & & \\
    & & & & & & & &\\
    \hline
    \end{tabular}
    \caption{Benchmark metrics for each model (BERT and LLMs) and each dataset. With respect to BERT models, the reported results reflect the use of weighted loss function. Iterative stratification was used to split the data. Results may vary due to stochastic processes on training, such as in weight initialization and token sampling in the case of LLMs. Specific LLMs versions: gpt-5 refers to \texttt{gpt-5-2025-08-07}; gpt-5-nano, \texttt{gpt-5-nano-2025-08-07}; gpt-5-mini, \texttt{gpt-5-mini-2025-08-07}; gpt-4.1, \texttt{gpt-4.1-2025-04-14}. Gemini models are stable models and have the same name in the API. \\ 
    {*} `minimal', `low', `medium' and `high' refer to the reasoning effort. \\ 
    {**} Inference with gpt-5 with high reasoning effort was aborted due to very long inference time and higher cost, even with optimized prompt.}
    \label{tab:benchmark}
\end{table*}

\begin{table*}
\centering
    \begin{tabular}{lcl}
    \hline
    \textbf{Methods} & \textbf{Macro F1-score ($\pm$ std)} & \textbf{Micro F1-score ($\pm$ std)}\\       
    \hline
    \textbf{SemClinBr} \\
    Random data split & 0.6567 ($\pm$ 0.0200) & 0.6904 ($\pm$ 0.0287) \\
    IS (Iterative Stratification) & 0.7139 ($\pm$ 0.0050) & 0.7646 ($\pm$ 0.0094) \\
    IS with weighted loss & 0.6672 ($\pm$ 0.0094) & 0.6773 ($\pm$ 0.0140) \\
    IS with weighted loss and clamped weights & 0.7118 ($\pm$ 0.0125) & 0.7604 ($\pm$ 0.0202) \\
    IS with OS (oversampling) & 0.7154($\pm$ 0.0117) & 0.7634 ($\pm$ 0.0073) \\
    \\
    \textbf{Breast cancer dataset} \\
    IS & 0.6928 ($\pm$ 0.0290) & 0.7301 ($\pm$ 0.0199) \\
    IS with weighted loss & 0.7283 ($\pm$ 0.0091) & 0.7591 ($\pm$ 0.0036) \\
    IS with weighted loss and clamped weights & 0.7340 ($\pm$ 0.0115) & 0.7634 ($\pm$ 0.0079) \\
    IS with OS (oversampling) & 0.7140 ($\pm$ 0.0110) & 0.7584 ($\pm$ 0.0106) \\
    \hline
    \end{tabular}
    \caption{Comparison of methods to deal with imbalanced classes using \texttt{mmBERT-base} model. For this comparison, both datasets were used with the IO data annotation scheme. Training for each data split method was repeated for 10 iterations. The mean and standard deviation (std) of the micro and macro F1-score are reported. Clamped weights, in this context, mean fixing the minimum weight for the classes at 1.0.}
    \label{tab:imbalancedexp}
\end{table*}

\begin{figure*}
  \includegraphics[width=0.38\linewidth]{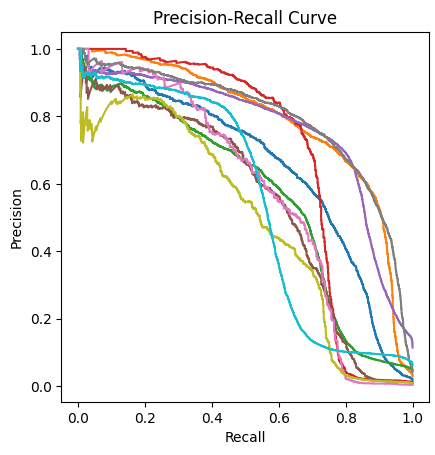} \hfill
  \includegraphics[width=0.60\linewidth]{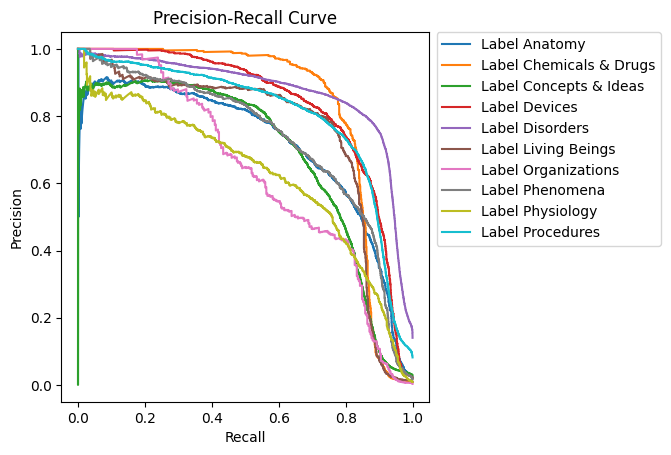}
  \caption {Left image: precision-recall curve using random data split. Right image: precision-recall curve using iterative stratification. The iterative stratification results in curves that better approximate the top-right corner of the plot, demonstrating better precision-recall trade-off.}
  \label{fig:randomvsiterative}
\end{figure*}

\subsection{Dealing with imbalanced classes}

Two data splitting strategies were compared: simple random split and iterative stratification. The impact of these methods was measured using the best-performing BERT method in this study: \texttt{mmBERT base}. 

In the SemClinBr dataset with the IO data annotation scheme, \texttt{mmBERT} achieved a maximum micro F1 score of 0.6904 ($\pm$ 0.0287) with a simple random split and 0.7646 ($\pm$ 0.0094) with iterative stratification. Detailed metrics are presented in \textbf{Table \ref{tab:imbalancedexp}}. The precision-recall curves are presented in \textbf{Figure \ref{fig:randomvsiterative}}.

Tests using weighted loss  and oversampling to deal with imbalanced classes were also explored. Due to the high frequency of certain classes, some computed weights were very small, limiting the model's capacity to learn those classes. In these cases, defining the minimum weight value as 1.0 was also explored. 

The results show that the best approach for dealing with class imbalance differs among datasets. In the SemClinBr, maximum performance was achieved through the iterative stratification of data alone. In the breast cancer dataset, it was the combination of iterative stratification with a weighted loss, with the minimum weight fixed at 1.0.

\section{Discussion}

\subsection{Key findings}

This study evaluated the performance of modern BERT models and LLMs on two clinical NER datasets: one public and the other private. BERT models, especially \texttt{mmBERT} \cite{mmbert} followed by BioBERTpt \cite{schneider-etal-2020-biobertpt}, demonstrated the best performance in the NER task across both datasets and achieved an F1-score of up to 0.7646. 

Several characteristics of \texttt{mmBERT} may have contributed to these results \cite{mmbert}. It was trained on about 3 trillion tokens, with hundreds of different languages represented. \texttt{BioBERTpt} \cite{schneider-etal-2020-biobertpt}, on the other hand, was fine-tuned on fewer than 50 million words in the Portuguese language. \texttt{mmBERT} incorporates language modeling optimizations from \texttt{ModernBERT} \cite{modernbert} and dynamically includes low-resource languages as the training phases progress, allowing for faster learning and better representation of these languages, even though fewer data is available for them. 

It is expected that additional in-domain pre-training improves model performance, as seen with BioBERTpt in the clinical context \cite{schneider-etal-2020-biobertpt}. Future work should explore fine-tuning mmBERT on biomedical data in the Portuguese language.

The possibility of data contamination cannot be excluded as a cause of the difference in performance between BioBERTpt and mmBERT in the SemClinBr dataset. mmBERT was trained on 3 trillion tokens and was published after the SemClinBr dataset was made public. The smaller difference in performance between both models in a private clinical dataset raises that concern even further. It was beyond the scope of this study to thoroughly investigate data contamination in the pretraining data of the evaluated BERT models, but this matter should be addressed in future studies.

The difference in performance between the IO and BIO annotation schemes, particularly in the macro F1-score, was greater in the \texttt{mmBERT} and \texttt{ModernBERT} models compared to \texttt{BioBERTpt} and \texttt{BERTimbau}. Since these results were obtained from the same datasets, it is reasonable to hypothesize that there are architectural aspects influencing how these models scale in larger label spaces. The BIO annotation scheme doubles the number of targeted classes by adding the beginning (B) tag to each class. Future work should explore these architectural differences and strategies to mitigate diminishing performance on less frequent classes.

\subsection{Strengths and limitations}

This study has several strengths. Our results established a new state-of-the-art for the SemClinBr dataset and demonstrated that these results can also be reproduced with private data. 

The private breast cancer dataset was annotated by a mixed team of clinicians and students, with disagreements resolved by consensus. This reflects a realistic, high-quality annotation pipeline that could be reproduced in diverse healthcare settings. With only 500 clinical notes, the authors were able to train a well performing NER model for oncological entities relevant to breast cancer care, demonstrating its feasibility. 

This study also systematically tested and quantified the effects of data iterative stratification, loss weighting, and data oversampling on multilabel imbalanced clinical data. The results pave the way for broader exploration of these techniques in clinical NLP tasks. With reproducibility as a cornerstone, all source code has been made available to the research community.

This study also has many limitations. Exploring LLMs' reasoning capabilities on NER tasks was beyond its scope. Time, cost, and data privacy concerns were among the reasons not to explore them further. No LLM fine-tuning was performed. The authors believe that small language models fine-tuned with biomedical data in the Portuguese language or prompt optimization strategies, such as GEPA \cite{agrawal2025gepareflectivepromptevolution}, may be a good option for clinical NER \cite{belcak2025smalllanguagemodelsfuture}. These strategies should be explored in future studies.

BERT models were not fine-tuned on additional biomedical data, which could influence the results. Future studies should investigate the impact of fine-tuning models on private health data on the performance of BERT models in public and private NER datasets. 

Although both datasets cover a broad range of topics and medical specialties, the complexity of healthcare is far greater than what this study was able to address. For this reason, the results may not apply to different medical specialties or to data from different regions.

Finally, this study did not include hypothesis testing to rigorously compare the different techniques for handling class imbalance, which limits the generalizability of the findings.

\section{Conclusion}

This study benchmarked modern BERT models and LLMs for clinical NER in the Portuguese language, using both public and private datasets. Among all tested models, mmBERT achieved the best performance, establishing a new state-of-the-art on the SemClinBr dataset and demonstrating strong generalization to a private breast-cancer corpus. The findings highlight the robustness of encoder-only architectures for clinical NER, especially when trained with multilingual strategies.

This work also contributes by systematically exploring strategies to handle multilabel class imbalance, such as iterative stratification, weighted loss, and oversampling. The results show that appropriate data-splitting strategies can substantially improve performance and that the use of oversampling and weighted loss may or may not help, depending on the data. The tests also showed that using weighted loss with a fixed  minimum weight of 1.0 for frequent classes preserves performance in those classes while also enhancing it in the less frequent ones. 

This study reinforces the feasibility of developing high-quality Portuguese clinical NER systems within real-world healthcare institutions, even with modest annotation efforts. Future research should explore domain-specific pre-training of multilingual models like mmBERT on Portuguese biomedical corpora, generalization across institutions and medical specialties, and  the use of refined small-language models on Portuguese clinical NER data.

%\section*{Acknowledgements}

%\section*{Author contributions}

%\section*{Disclosure}

\bibliography{custom}

%\appendix
\end{document}